%% file: ArxivSubmission2027.tex
\documentclass[letterpaper]{article}
\usepackage[preprint]{aaai2027}
\usepackage[hyphens]{url}
\usepackage{graphicx}
\usepackage{natbib}
\usepackage{caption}
\usepackage{booktabs}
\usepackage{amsmath,amssymb}
\usepackage{array}
\usepackage{multirow}
\usepackage{placeins}
\usepackage{float}
\makeatletter
\AtBeginEnvironment{thebibliography}{\footnotesize\setlength{\itemsep}{0.12em}\setlength{\parskip}{0pt}\setlength{\parsep}{0pt}}
\makeatother
\newcommand{\method}{\textsc{MOSH-WM}}
\newcommand{\savi}{\textsc{SAVi}}
\newcommand{\slotformer}{\textsc{SlotFormer}}

\title{MOSH-WM: Mask-Grounded Soft-Hamiltonian Dynamics for Object-Centric World Models}
\author{Zhekai Wang\textsuperscript{1*}, Haoxiang Huang\textsuperscript{2*}, Xiang Liu\textsuperscript{3}, Zhikang Chen\textsuperscript{3}, Yueqing Sun, Qi Gu, Shiji Zhou\textsuperscript{4}, Miao Liu\textsuperscript{3}, Sen Cui\textsuperscript{3 \dag}}
\affiliations{\textsuperscript{1}Beijing Institute of Technology\\
\textsuperscript{2}University of Science and Technology of China\\
\textsuperscript{3}Tsinghua University\\
\textsuperscript{4}Beihang University\\
\textsuperscript{*}Equal contribution. \quad \textsuperscript{\dag}Corresponding authors. 
}

\begin{document}
\maketitle

\begin{abstract}
Object-centric world models forecast future videos by evolving a set of entity slots, but the variables receiving dynamics supervision are often unconstrained visual features. We introduce \method{}, a mask-grounded soft-Hamiltonian world model that makes its position-like state explicitly depend on slot-owned image support. A frozen video-slot encoder produces slots and masks; spatial moments of mask-owned support form a canonical state $Q$, temporal differences form $P$, and a learned energy supplies a soft directional bias to a bounded learned increment. Decoder-relevant appearance and identity are stored separately in a causal visual context. A gated composer and bounded residual then combine this context with the propagated phase state to reconstruct decoder-compatible slots. On OBJ3D, given six observed frames and evaluated over the following 30 frames, \method{} reduces LPIPS by 25.0\% and spatial MSE by 33.7\% relative to the strongest object-centric baseline. On CLEVRER, given six observed frames and evaluated over the following ten frames, the corresponding reductions are 14.5\% and 18.7\%. Horizon-resolved visual and object-state measurements show that the complete model accumulates error more slowly throughout the 30-frame closed-loop rollout. Project page: \url{https://github.com/moshwm-anon/-moshwm-anon.github.io}.
\end{abstract}

\section{Introduction}
The physical world evolves through regularities that govern object motion and interaction. A useful visual world model should maintain these regularities when its predictions are repeatedly fed back. Object-centric learning provides a natural representation by decomposing a scene into entity slots \cite{slotattention,kipf2022savi}. Existing slot-based predictors evolve these slots autoregressively and decode them into future frames \cite{slotformer,ocvp}, yielding strong visual forecasts on many benchmarks.

\begin{figure}[!h]
\centering
\includegraphics[width=.96\columnwidth]{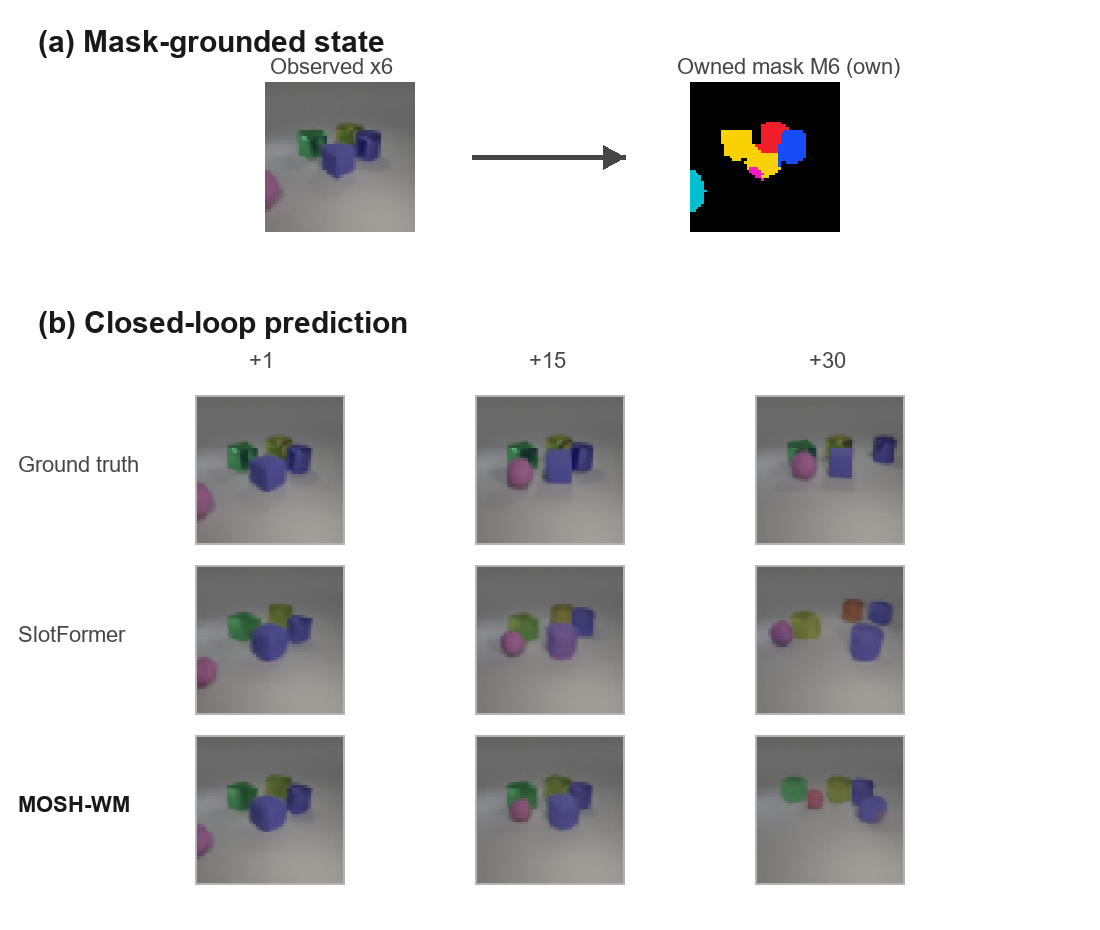}
\caption{Mask grounding and long-horizon prediction on OBJ3D. Owned masks expose object support. SlotFormer matches the immediate future but drifts over closed-loop rollout; \method{} better preserves object identity and scene layout.}
\label{fig:task-teaser}
\end{figure}

Most object-centric world models optimize slots for grouping and reconstruction, then fit video trajectories with a transition over these visual features. The slots can entangle geometry, appearance, identity, and decoder-specific semantics, leaving long closed-loop rollouts weakly constrained. Figure~\ref{fig:task-teaser} illustrates the resulting gap: a strong slot predictor matches the first future frame, but its object layout and appearance drift over a 30-step rollout. A useful physical state should instead be grounded in where each object is supported while retaining separate visual information for decoding.

Hamiltonian models organize dynamics through generalized coordinates $Q$, generalized momenta $P$, and an energy function \cite{greydanus2019hamiltonian,hgn,slotpi}. Applying this structure directly to arbitrary visual slots remains problematic: the resulting coordinates may not describe object geometry, while a compact physical state may omit information needed by the decoder. The model must therefore extract a geometric phase state $(Q,P)$ and preserve the visual factors needed for decoding. Hamiltonian guidance acts on the former; appearance and identity follow a separate contextual pathway.

\method{} implements this factorization with a frozen \savi{} interface. It assigns each pixel to its owning slot and pools fixed spatial basis functions over that support to construct a position-like $Q$; raw finite differences yield the momentum-like $P$. The extractor uses owned masks and image-coordinate bases, excluding appearance-bearing slot vectors. A soft-Hamiltonian module propagates $(Q,P)$ by blending energy-gradient directions with a bounded learned field. The Hamiltonian term serves as a structural prior, without claiming exact system identification. In parallel, causal slot history retains appearance and identity. A gated composer with bounded residual correction then produces slots for the frozen decoder.

\noindent\textbf{Contributions.} Our contributions are threefold:
\begin{itemize}
    \item A \emph{mask-grounded phase-state extractor} computes $Q$ from per-slot owned masks and fixed spatial bases, with raw temporal differences defining $P$. Appearance-bearing slot features are excluded from this dynamics input.
    \item A \emph{soft-Hamiltonian closed-loop dynamics module} combines energy-gradient directions with a bounded learned field and directly propagates $(Q,P)$.
    \item A \emph{decoder reconciliation interface} uses causal context, gated slot composition, and bounded residual compensation to recover decoder-compatible slots.
\end{itemize}

\section{Related Work}
\paragraph{Object-centric video world models.} Object-centric learning decomposes a scene into exchangeable entities through sequential attention, iterative inference, structured generative models, or slot attention \cite{air,iodine,monet,space,genesis,genesisv2,slotattention}. Earlier scene and control models also use entity abstraction or compositional rendering \cite{gqn,op3}. Video models such as SCALOR and \savi{} extend this interface with temporal correspondence and persistent object slots \cite{scalor,kipf2022savi}; G-SWM, Visual Interaction Networks, Neural Physics Engines, interaction networks, and neural relational inference explicitly model object interactions \cite{gswm,vin,npe,interactionnetworks,nri}. Recent predictors, including Object-Centric Video Prediction, SlotFormer, and slot-mixing architectures, autoregressively evolve visual slots to generate future frames \cite{ocvp,slotformer,slotmixer}. Related representation learners study structured world models, transformer-based slots, and real-world scaling \cite{cswm,steve,dinosaur}. These methods establish slots as an effective prediction interface, but their transitions generally act on holistic visual features; geometry, appearance, identity, and decoder semantics need not be separated. \method{} instead uses video slots primarily for perception and contextual decoding, while deriving its canonical dynamics state from slot-owned mask support.

\paragraph{Hamiltonian and physics-informed dynamics.} Hamiltonian neural networks parameterize an energy function and obtain dynamics from phase-space gradients \cite{greydanus2019hamiltonian}; Lagrangian, symplectic, continuous-time, and graph-based extensions strengthen geometric or interaction structure \cite{cranmer2020lagrangian,deeplagrangian,symplectic,ode,latentode,gns,gnns}. Hamiltonian Generative Networks infer latent phase variables from image sequences \cite{hgn,vintegrator}. At the object-centric level, SlotPi uses attention over slots to estimate generalized coordinates and momenta before combining a physical module with spatiotemporal reasoning \cite{slotpi}; HaM-World factorizes planner latents into $(q,p,c)$ and applies a soft-Hamiltonian prior to the canonical subspace \cite{hamworld}. In contrast, \method{} grounds $Q$ in owned image support and obtains $P$ by finite differences, then uses the Hamiltonian field as a soft closed-loop bias while reserving visual context for decoder compatibility.

\paragraph{Latent and pixel-space world models.} Latent world models, from World Models and PlaNet to Dreamer and MuZero, learn compact rollouts for planning or behavior learning from pixels \cite{worldmodels,planet,dreamer,dreamerv2,muzero}. Stochastic video models separate motion uncertainty from visual content \cite{svg,tsp}, while pixel-space predictors such as PredRNN and SimVP directly optimize future-frame quality \cite{predrnn,simvp}. These are useful visual controls, but they do not expose temporally aligned object states. Physical-reasoning benchmarks and relational predictors such as PHYRE, Physion, CLEVRER, and RPIN emphasize interactions, interventions, or counterfactual structure \cite{phyre,physion,clevrer,rpin}. Our goal is complementary: to study whether a mask-grounded, softly Hamiltonian state can make closed-loop object-centric prediction more structurally interpretable while retaining a decoder-compatible visual pathway.

\section{Method}
\subsection{Overview}
\begin{figure*}[t]
\centering
\includegraphics[width=\textwidth]{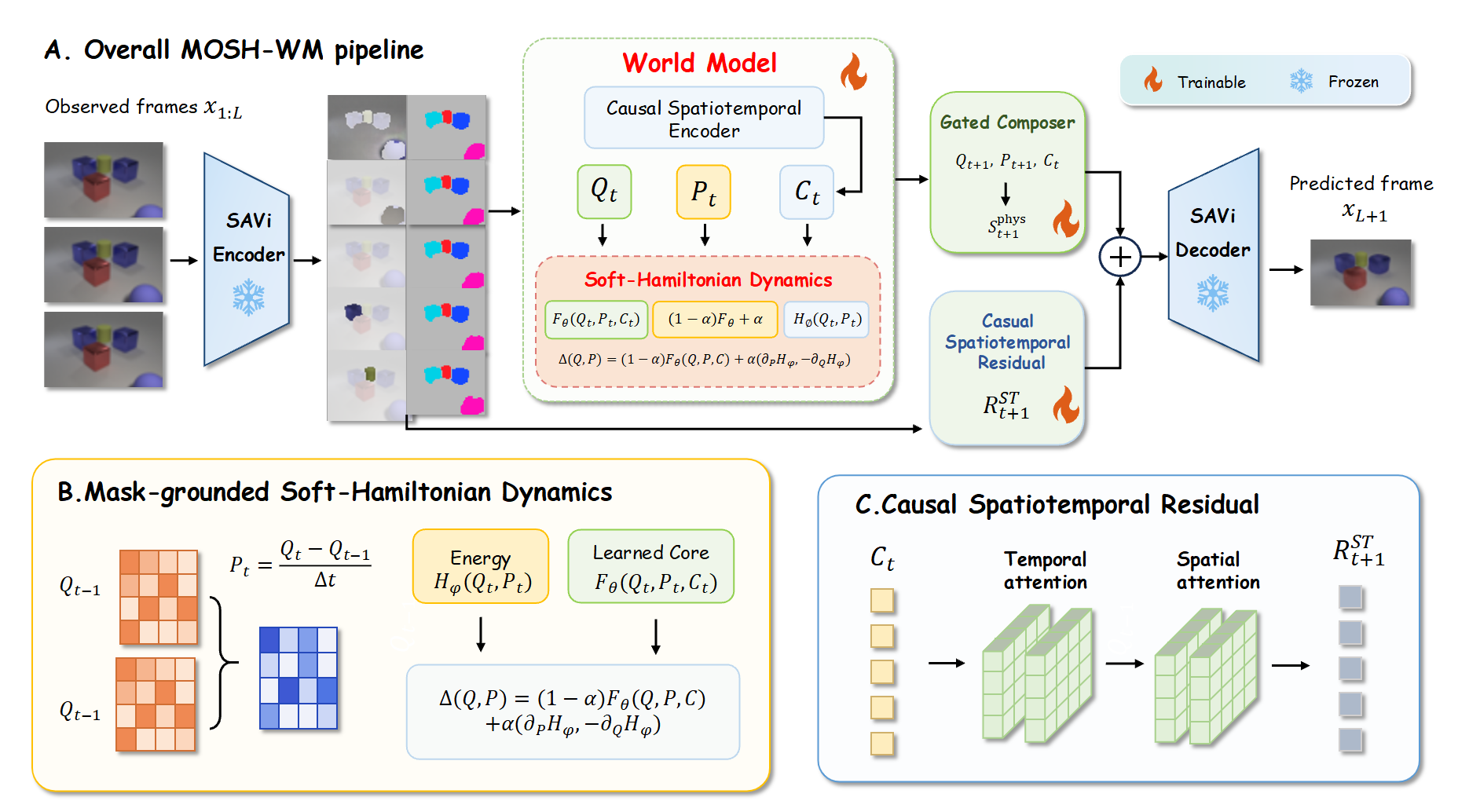}
\caption{\method{} pipeline. A frozen \savi{} interface supplies visual slots and decoded masks. Mask-owned pooling extracts $Q$, raw temporal differences provide $P$, and causal slots provide $C$. The \textsc{SoftHamiltonian} update propagates $(Q,P)$; a gated composer and bounded spatiotemporal residual return to decoder-compatible slot space. Dashed arrows denote direct state propagation and autoregressive context update.}
\label{fig:pipeline}
\end{figure*}

Given $L$ observed frames $X_{1:L}$, the frozen \savi{} interface \cite{kipf2022savi} returns visual slots $S_{1:L}$ and soft masks $M_{1:L}$. \method{} maintains a canonical phase state $(Q_t,P_t)$ and a causal visual context $C_t$. The phase state is used exclusively by the Soft-Hamiltonian transition, whereas $C_t$ retains appearance and identity information for decoding. Attention-based exchange follows the general token-interaction pattern of Transformers and Perceiver-style modules \cite{transformer,perceiver}. Thus, the model does not require one visual slot to act simultaneously as a physical coordinate and a complete decoder input. Figure~\ref{fig:pipeline} summarizes the two closed loops: $(Q,P)$ are propagated directly, while predicted slots update the causal context.

\subsection{Mask-Grounded Phase-State Factorization}
Slots are optimized for grouping and reconstruction and may therefore entangle geometry with appearance and semantics \cite{slotattention,kipf2022savi}. We retain the support owned by slot $n$ through winner-take-all assignment, $M^{\mathrm{own}}_{t,n}(u,v)=M_{t,n}(u,v)\mathbb I[n=\arg\max_j M_{t,j}(u,v)]$. Let $b$ be a fixed spatial basis and let $\operatorname{Pool}$ be normalized mask-weighted pooling. The position-like coordinate is
\begin{equation}
Q_{t,n}=f_Q\left(\operatorname{Pool}(M^{\mathrm{own}}_{t,n},b)\right).
\label{eq:qstate}
\end{equation}
The lightweight map $f_Q$ contains a basis projector, a slot embedding, and one attention-MLP block. Its inputs are mask support and fixed spatial bases, which separate the canonical coordinate from direct slot appearance. Its momentum-like companion is the temporal finite difference,
\begin{equation}
P_t=\frac{Q_t-Q_{t-1}}{\Delta t}.
\label{eq:pstate}
\end{equation}
In parallel, a causal spatiotemporal encoder reads slot history and produces $C_t=\mathcal F_{\mathrm{ctx}}(S_{\leq t})$. Context never enters the phase-state extractor; it is reserved for the decoder-compatible pathway below.

\subsection{Soft-Hamiltonian Closed-Loop Dynamics}
A fully learned transition can fit local visual correlations without a preferred physical direction, while a strict Hamiltonian model can be too restrictive for imperfect visual observations. We therefore use a soft physical bias. The energy head receives only the canonical pair,
\begin{equation}
H_t=H_{\vartheta}(Q_t,P_t).
\label{eq:energy}
\end{equation}
A learned core conditioned on $[Q_t,P_t]$ and $C_t$ predicts increments $(D_t^Q,D_t^P)$. We blend these increments with the energy-gradient directions:
\begin{equation}
Q_{t+1}=Q_t+(1-\alpha)D_t^Q+\alpha\operatorname{clip}_{\gamma}(\nabla_{P_t}H_t).
\label{eq:qupdate}
\end{equation}
\begin{equation}
P_{t+1}=P_t+(1-\alpha)D_t^P-\alpha\operatorname{clip}_{\gamma}(\nabla_{Q_t}H_t).
\label{eq:pupdate}
\end{equation}
The learned path captures residual dynamics, whereas the gradient path organizes phase evolution. These are forward-Euler Soft-Hamiltonian updates, not a symplectic solver or a claim that $H_t$ is ground-truth mechanical energy. During rollout, the model directly sets $(Q_t,P_t)\leftarrow(Q_{t+1},P_{t+1})$; the update scale and gradient clipping are specified with the learning objective below.

\subsection{Physics--Appearance Reconciliation}
The phase state intentionally omits decoder-relevant visual information. After the physical update, we use a gated composer to combine the normalized phase displacement with $C_t$, yielding a physical slot proposal $S_{t+1}^{\mathrm{phys}}$. A causal reasoning module then provides a small decoder-space correction:
\begin{equation}
\widehat S_{t+1}=S_{t+1}^{\mathrm{phys}}+\beta\left(\mathcal R(S_t,S_{\leq t})-S_t\right).
\label{eq:reconcile}
\end{equation}
The composer applies learned projections to normalized phase displacement and $C_t$, predicts a sigmoid gate, and blends the two projected features before mapping them to slot space. The frozen decoder produces $\widehat X_{t+1}=\mathcal D(\widehat S_{t+1})$, and $\widehat S_{t+1}$ is appended to the history used for $C_{t+1}$. Hence, visual context can improve decoding without being fed back into the mask-grounded phase extractor.

\subsection{Learning Objective}
We freeze the \savi{} encoder and decoder, encode $L=6$ frames, and unroll $K$ future steps according to the training curriculum. The complete objective is
\begin{align}
\mathcal L={}&\lambda_s\mathcal L_{\mathrm{slot}}+\lambda_x\mathcal L_{\mathrm{img}}
+\lambda_h\mathcal L_{\mathrm{ham}}+\lambda_e\mathcal L_{\mathrm{energy}}\nonumber\\
&+\lambda_c\mathcal L_{\mathrm{ctr}}+\lambda_r\mathcal L_{\mathrm{res}}.
\label{eq:objective}
\end{align}
$\mathcal L_{\mathrm{slot}}$ keeps predicted slots compatible with the frozen observer and decoder, while $\mathcal L_{\mathrm{img}}$ supervises decoded appearance using pixel, multiscale, and edge consistency. $\mathcal L_{\mathrm{ham}}$ aligns the learned phase increment with the Hamiltonian gradient direction, and $\mathcal L_{\mathrm{energy}}$ discourages abrupt local and rollout-level energy drift. $\mathcal L_{\mathrm{ctr}}$ directly supervises mask-derived object trajectories. $\mathcal L_{\mathrm{res}}$ bounds the visual correction to the phase-composed slot. On OBJ3D, $(\lambda_s,\lambda_x,\lambda_h,\lambda_e,\lambda_c,\lambda_r)=(1,1,0.05,0.01,0.5,0.5)$. The rollout component inside $\mathcal L_{\mathrm{energy}}$ has weight 0.25; $\alpha=0.5$, Hamiltonian gradients are clipped at 5, and the residual gain increases to 0.04. Further definitions and schedules are given in the supplement.

\section{Experiments}
We organize the experiments around four questions: (1) does the mask-grounded phase interface preserve decoder-compatible appearance; (2) does \method{} improve visual quality and object dynamics on the primary OBJ3D and CLEVRER benchmarks; (3) does the advantage persist under long-horizon feedback and component removal; and (4) does the factorization transfer to a different frozen observer on Physion? The section follows that order so the evidence builds from interface fidelity to benchmark performance, stability, ablation, and transfer.

\subsection{Experimental Setup}

\paragraph{Datasets and protocols.}
OBJ3D \cite{gswm} contains controlled multi-object motion in $64\times64$ videos. The primary protocol observes six frames and evaluates the next 30 frames, with the \method{} curriculum ending at the same horizon. CLEVRER \cite{clevrer}, built on the compositional CLEVR scene family \cite{clevr}, contains smaller objects, collisions, occlusions, and multi-object interactions; six observations are followed by ten predictions. Physion \cite{physion} contains rendered physical interactions across Collide, Contain, Dominoes, Drape, Drop, Link, Roll, and Support scenarios. Following the SlotFormer protocol, each $128\times128$ video is truncated to 150 source frames and sampled every three frames. STEVE produces six 192-dimensional slots; MOSH observes 15 sampled frames and predicts the remaining 35 without target feedback.

\paragraph{Implementation and evaluation protocol.}
Methods share the observation window, test partition, evaluator, and metrics within each protocol, while the \method{} observer and decoder remain frozen. On OBJ3D, the world model uses six 128-dimensional slots, four attention heads, one $Q$ block, one energy block, two context blocks, two reasoning blocks, MLP expansion ratio 2, and composer width 256. We train with batch size 8, Adam at learning rate $10^{-5}$, 5\% warm-up followed by cosine decay, zero weight decay, gradient clipping at 1, and 8,000 steps with a 10/15/20/24/30-frame rollout curriculum; validation Slot@30 selects the checkpoint without using test data.
CLEVRER uses seven slots, batch size 20, learning rate $8\times10^{-6}$, and 2,400 constrained-stage steps with a 6/8/10-frame curriculum. For Physion, STEVE is adapted on 250 videos per scenario and selected by validation token reconstruction; MOSH is selected independently by validation slot MSE@30.
All methods receive the same six observed frames and predict 30 future frames without target feedback on OBJ3D. LPIPS is averaged per predicted frame; PSNR and SSIM use images mapped from $[-1,1]$ to $[0,1]$. Object-state metrics re-encode predictions and targets with the frozen observer: centroid error measures normalized foreground displacement, velocity error compares first differences, pair-distance error compares foreground-object pairs, and Slot@30 excludes the largest-area background slot. Physion reports horizon-resolved slot MSE on all 512 test videos and decoded metrics on a scenario-balanced 64-video subset. The transfer studies are single validation-selected runs, so we treat them as matched transfer evidence rather than multi-seed comparisons.
\begin{figure*}[!t]
\centering
\includegraphics[width=.98\textwidth]{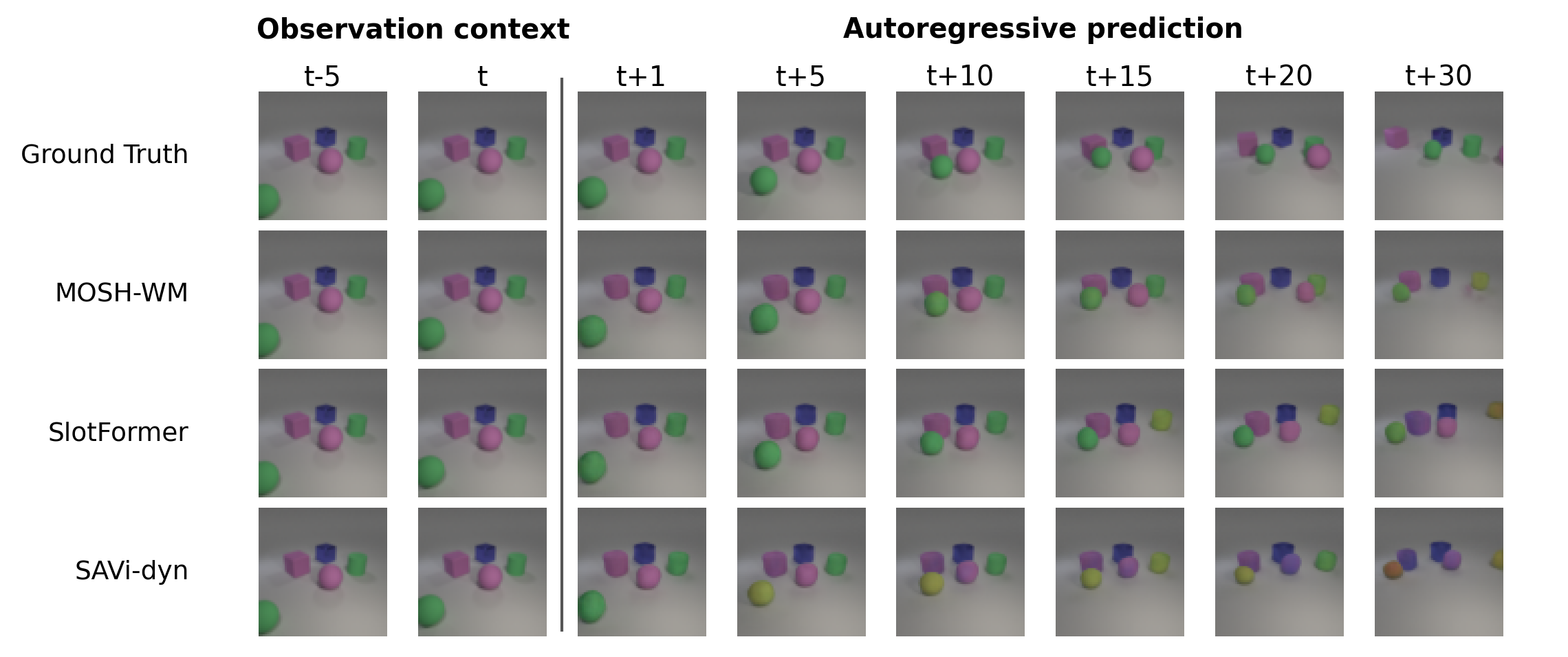}
\caption{Long-sequence prediction on OBJ3D. Two observation frames and six future frames are shown at a larger scale than a full-frame strip. All rows use the same six-frame context and independently exported 30-frame closed-loop predictions. Under repeated feedback, \method{} better retains object identity, color, and relative layout.}
\label{fig:qualitative}
\end{figure*}

\subsection{Evaluation on Video Prediction}

\paragraph{Baselines.}
The primary object-centric baselines are \slotformer{} \cite{slotformer} and SAVi-dyn \cite{kipf2022savi}. They share the frozen SAVi observer and decoder with \method{} on OBJ3D and CLEVRER. PredRNN \cite{predrnn} provides a pixel-space visual baseline, while object-state metrics apply only to methods with aligned slots.

\paragraph{Evaluation metrics.}
We report LPIPS \cite{lpips}, PSNR, SSIM \cite{ssim}, and spatial MSE for decoded videos; CLEVRER additionally uses foreground MSE. OBJ3D dynamics use centroid ADE/FDE, velocity error, pairwise-distance error, and observer-slot MSE. Physion uses horizon-resolved slot MSE on all 512 test videos and STEVE-decoded PSNR, SSIM, and LPIPS on a scenario-balanced 64-video test subset.

\paragraph{Object-centric interface.}

\begin{figure}[!t]
\centering
\includegraphics[width=\columnwidth]{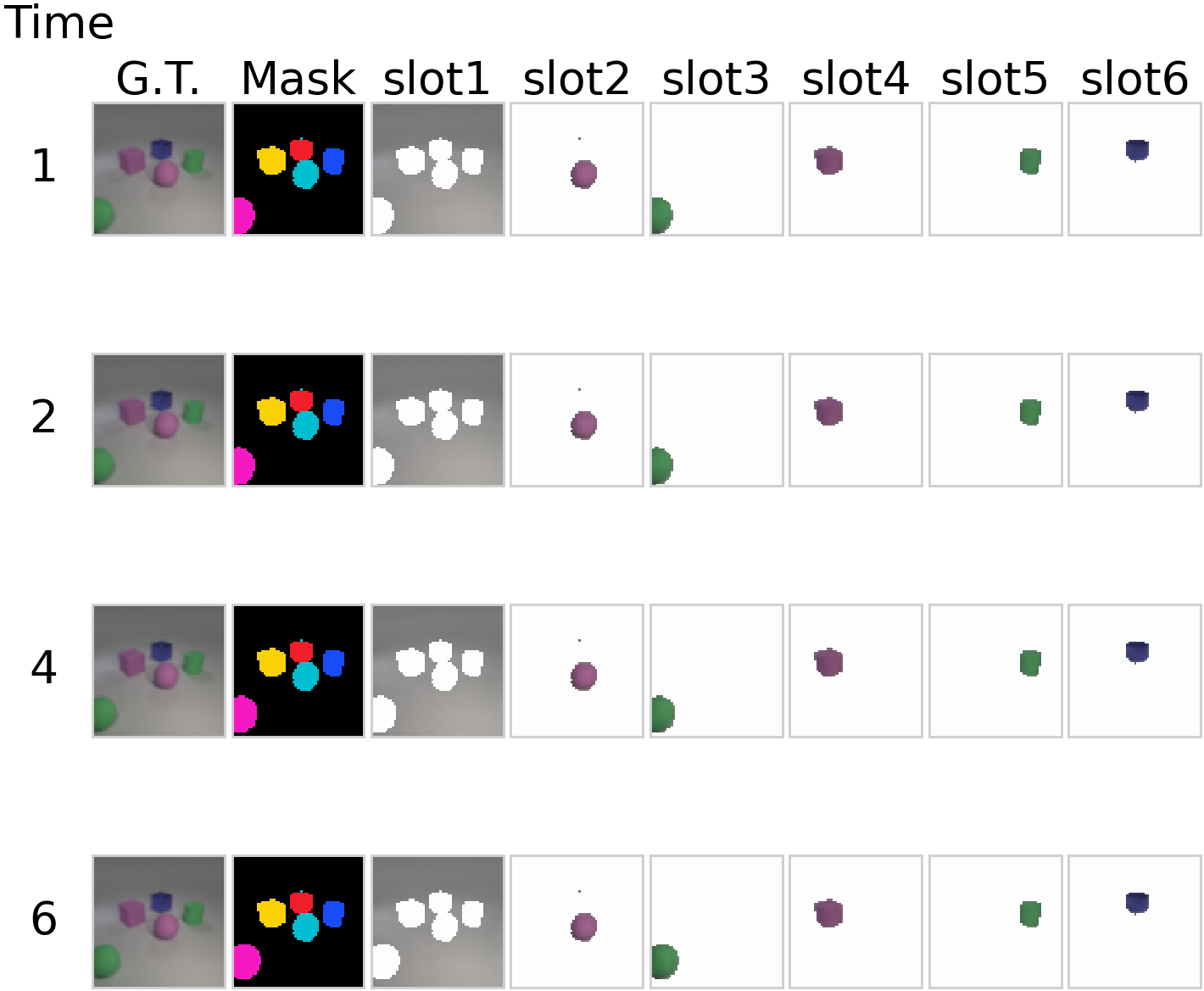}
\caption{SlotPi-style scene decomposition by the frozen SAVi observer on OBJ3D. Each row is one observation time. The mask uses a black background and mutually exclusive colors; slots 1--6 are independently decoded RGB components on near-white backgrounds. Fixed columns expose temporal slot consistency and the support used to construct $Q$.}
\label{fig:savi-perception}
\end{figure}

Before comparing dynamics, we verify the interface from which \method{} constructs $(Q,P)$. Figure~\ref{fig:savi-perception} follows the scene-decomposition visualization of SlotPi: each row is one observation time, and each sequence is arranged as the ground-truth frame, a mutually exclusive color mask, and six independently decoded RGB slots on white backgrounds. Slot columns are fixed over time. Foreground objects remain assigned to consistent slots across the observation window, while the background component is separated from the object slots. This owned support is pooled by Equation~\ref{eq:qstate}. The RGB slot paired with each mask supplies the appearance and identity features retained in the causal context.

\begin{table}[!t]
\centering\small
\setlength{\tabcolsep}{5.5pt}
\begin{tabular}{lrrrr}
\toprule
Reconstruction & LPIPS $\downarrow$ & PSNR $\uparrow$ & SSIM $\uparrow$ & MSE $\downarrow$\\
\midrule
Frozen SAVi & \textbf{0.0286} & \textbf{41.05} & \textbf{0.9826} & \textbf{0.376}\\
Phase composer & 0.0288 & 40.81 & 0.9820 & 0.395\\
\bottomrule
\end{tabular}
\caption{OBJ3D reconstruction quality before rollout. The phase-to-slot composer remains close to the frozen-interface upper bound.}
\label{tab:savi-reconstruction}
\end{table}

Table~\ref{tab:savi-reconstruction} quantifies the same interface. Mapping the mask-grounded phase representation back to decoder-compatible slots changes LPIPS by only 0.0002 and PSNR by 0.24 dB relative to direct frozen-SAVi reconstruction. This is the first evidence that the phase-state factorization preserves decoder-compatible appearance before rollout begins.

\paragraph{Visual prediction results.}

\begin{table*}[!t]
\centering\small
\begin{minipage}[t]{.485\textwidth}
\centering
\textbf{(a) OBJ3D visual prediction (6 observed $\rightarrow$ 30 predicted)}\\[2pt]
\setlength{\tabcolsep}{2.4pt}
\begin{tabular}{@{}lrrrr@{}}
\toprule
Method & LPIPS $\downarrow$ & PSNR $\uparrow$ & SSIM $\uparrow$ & MSE $\downarrow$\\
\midrule
\method{} & \textbf{0.0858} & \textbf{31.02} & \textbf{0.8916} & \textbf{4.718}\\
\slotformer{} & 0.1144 & 28.59 & 0.8503 & 7.119\\
SAVi-dyn & 0.1197 & 28.35 & 0.8522 & 7.368\\
PredRNN & 0.1657 & 30.98 & 0.8823 & 5.258\\
\bottomrule
\end{tabular}
\end{minipage}\hfill
\begin{minipage}[t]{.485\textwidth}
\centering
\textbf{(b) CLEVRER visual prediction (6 observed $\rightarrow$ 10 predicted)}\\[2pt]
\setlength{\tabcolsep}{1.7pt}
\begin{tabular}{@{}lrrrrr@{}}
\toprule
Method & LPIPS $\downarrow$ & PSNR $\uparrow$ & SSIM $\uparrow$ & MSE $\downarrow$ & FG-MSE $\downarrow$\\
\midrule
\method{} & \textbf{0.3192} & \textbf{26.60} & \textbf{0.8375} & \textbf{10.068} & \textbf{0.00996}\\
\slotformer{} & 0.3814 & 25.58 & 0.7953 & 12.379 & 0.01223\\
SAVi-dyn & 0.3736 & 25.51 & 0.7944 & 12.558 & 0.01240\\
\bottomrule
\end{tabular}
\end{minipage}

\vspace{5pt}
\textbf{(c) OBJ3D dynamics measured through the shared frozen observer}\\[2pt]
\setlength{\tabcolsep}{10pt}
\begin{tabular}{lrrrrr}
\toprule
Method & ADE $\downarrow$ & FDE $\downarrow$ & Velocity $\downarrow$ & Pair distance $\downarrow$ & Slot MSE $\downarrow$\\
\midrule
\method{} & \textbf{.00473} & \textbf{.00957} & \textbf{.00127} & \textbf{.00592} & \textbf{.00795}\\
\slotformer{} & .00559 & .01057 & .00160 & .00696 & .01230\\
SAVi-dyn & .00597 & .01098 & .00136 & .00717 & .01173\\
\bottomrule
\end{tabular}
\caption{Primary results under shared protocols. Bold denotes the best result in each dataset block. OBJ3D uses 6-to-30 prediction and CLEVRER uses 6-to-10 prediction. Object-state metrics use the shared frozen observer.}
\label{tab:benchmark-main}
\end{table*}

\paragraph{OBJ3D visual prediction.}
Table~\ref{tab:benchmark-main}(a) shows that the final 6-to-30 model improves perceptual, distortion, and structural metrics over both reported object-centric baselines. Relative to the strongest object-centric baseline, LPIPS and spatial MSE fall by 25.0\% and 33.7\%, PSNR improves by 2.43 dB, and SSIM reaches 0.8916. \method{} also outperforms the reported pixel-space PredRNN baseline on all four visual metrics. Together with the reconstruction test above, this closes the chain from interface fidelity to end-to-end forecast quality.

\paragraph{OBJ3D object dynamics.}
The common-observer measurements in Table~\ref{tab:benchmark-main}(c) show corresponding object-level gains. \method{} lowers centroid ADE by 15.4\%, FDE by 9.5\%, velocity error by 6.6\%, pairwise-distance error by 14.9\%, and slot MSE by 32.2\% relative to the strongest baseline in each column. The lower pairwise-distance error reflects better preservation of the scene configuration. Image-space and observer-space metrics thus agree on more accurate state evolution.

\paragraph{CLEVRER.}
Table~\ref{tab:benchmark-main}(b) evaluates transfer to collisions and occlusion. Under the shared local protocol, \method{} improves all five visual metrics over \slotformer{} and SAVi-dyn. The time-boxed single-run comparison supports transfer within this matched setup.

\paragraph{Qualitative behavior.}
Figure~\ref{fig:qualitative} displays early and late prediction times jointly. The three methods remain close immediately after the observation window. At later steps, the baselines increasingly alter object color, shape, or relative position, while \method{} stays closer to the target. The widening gap indicates slower closed-loop drift.

\paragraph{Long-horizon error accumulation.}
\label{sec:long-horizon}

\begin{table}[!t]
\centering\small
\setlength{\tabcolsep}{2.6pt}
\begin{tabular}{@{}lrrr@{}}
\toprule
Method & Mean@10 & Mean@30 & Final@30\\
\midrule
\method{} & \textbf{.0381} & \textbf{.0858} & \textbf{.1744}\\
\slotformer{} & .0540 & .1144 & .2063\\
SAVi-dyn & .0559 & .1197 & .2065\\
\bottomrule
\end{tabular}
\caption{OBJ3D mean LPIPS up to increasing horizons and LPIPS at the final predicted frame.}
\label{tab:horizon-analysis}
\end{table}

Table~\ref{tab:horizon-analysis} evaluates the same checkpoints in continuous rollouts. \method{} leads over the first ten predicted frames, and its absolute gap to \slotformer{} grows from 0.0159 at Mean@10 to 0.0286 at Mean@30. Final-frame LPIPS is also lower (0.1744 versus 0.2063), confirming that the advantage does not vanish at the rollout endpoint.

\begin{figure}[!t]
\centering
\includegraphics[width=\columnwidth]{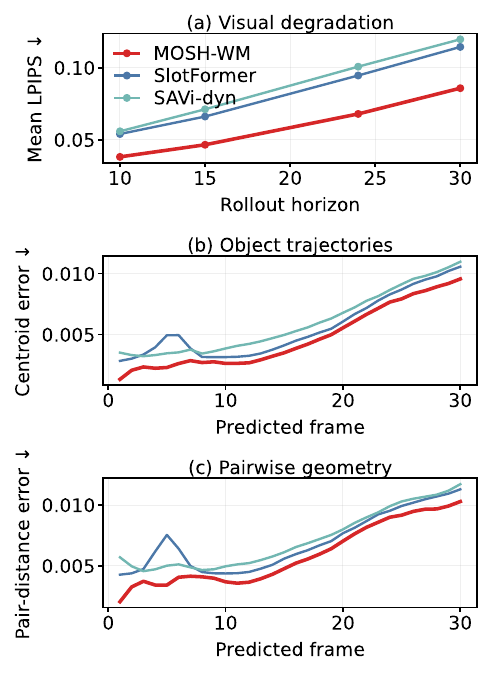}
\caption{OBJ3D closed-loop error accumulation. (a) Mean LPIPS over increasing horizons. (b) Per-frame centroid error. (c) Per-frame pairwise-distance error. All curves use the same six observations.}
\label{fig:horizon}
\end{figure}

Figure~\ref{fig:horizon} shows slower visual-error growth for \method{}. Its centroid and pairwise-geometry curves also remain below the competing object-centric transitions throughout the rollout. The gain spans decoded frames, individual trajectories, and object relations. Direct propagation of a mask-grounded state keeps later predictions anchored to an explicit geometric path.

\paragraph{Distortion over time.}
The same trend holds in pixel space: \method{} approximately halves mean spatial MSE over the first ten predictions and retains a 33.7\% advantage at Mean@30 relative to \slotformer{}. Table~\ref{tab:horizon-analysis} and Figure~\ref{fig:horizon} confirm the gain across metrics and rollout horizons, so the long-horizon claim is supported by both image and object-state measurements.

\subsection{Ablation Study}
\paragraph{OBJ3D components.}
Table~\ref{tab:component-ablation} evaluates every variant with the same six observations and 30-step closed-loop rollout. The trained variants retain their respective optimization protocols; the comparison here standardizes inference rather than imposing an identical training budget. The two additional $\alpha$ settings are independently fine-tuned for 2,000 steps from the long-rollout model rather than applied only at inference time. This makes the ablation about component sensitivity rather than a pure post hoc switch of modules.

The complete model gives the best visual and slot-space results. Removing physics nearly doubles LPIPS and more than triples Slot@30, while removing causal context also causes a large visual degradation. Physics-only prediction is substantially stronger than residual-only prediction, but remains behind their learned combination. Re-encoding the propagated state also increases long-horizon slot error. Among the added mixing coefficients, $\alpha=0.2$ is slightly better than $\alpha=0.1$, but neither improves over the default $\alpha=0.5$ model. The ablation therefore supports both submodules and their coupling.

\begin{table}[!t]
\centering
\small
\setlength{\tabcolsep}{3.2pt}
\begin{tabular}{@{}lrrr@{}}
\toprule
Variant & LPIPS $\downarrow$ & MSE $\downarrow$ & Slot@30 $\downarrow$\\
\midrule
\method{} ($\alpha=0.5$) & \textbf{0.0858} & \textbf{4.718} & \textbf{0.02608}\\
\quad residual only & 0.1665 & 8.468 & 0.08405\\
\quad physics only & 0.1037 & 5.840 & 0.04212\\
\quad w/o context & 0.1437 & 8.595 & 0.04799\\
\quad re-encode state & 0.1122 & 6.259 & 0.06985\\
\quad $\alpha=0.1$ & 0.0964 & 5.154 & 0.02970\\
\quad $\alpha=0.2$ & 0.0961 & 5.140 & 0.02939\\
\bottomrule
\end{tabular}
\caption{OBJ3D component and mixing-coefficient ablation. All entries use the same six-input, 30-prediction evaluator. Bold identifies each column minimum.}
\label{tab:component-ablation}
\end{table}

Additional capacity ablations clarify whether the gain depends on a narrowly tuned transition width. Under the shared short-budget protocol, two attention blocks achieve LPIPS/MSE/Slot@30 of $0.0873/4.789/0.02755$, compared with $0.1863/9.945/0.06165$ for one block and $0.1251/6.856/0.03984$ for three blocks. The same pattern appears in the MLP sweep: the default expansion ratio of two gives $0.0873/4.789/0.02755$, whereas ratios one and four degrade to $0.2662/12.266/0.07734$ and $0.2193/10.887/0.06511$. Thus, the reported improvement is not explained by simply increasing capacity; the two-block configuration provides the best balance between expressivity and rollout stability under the shared budget.

\begin{figure}[!t]
\centering
\includegraphics[width=\columnwidth]{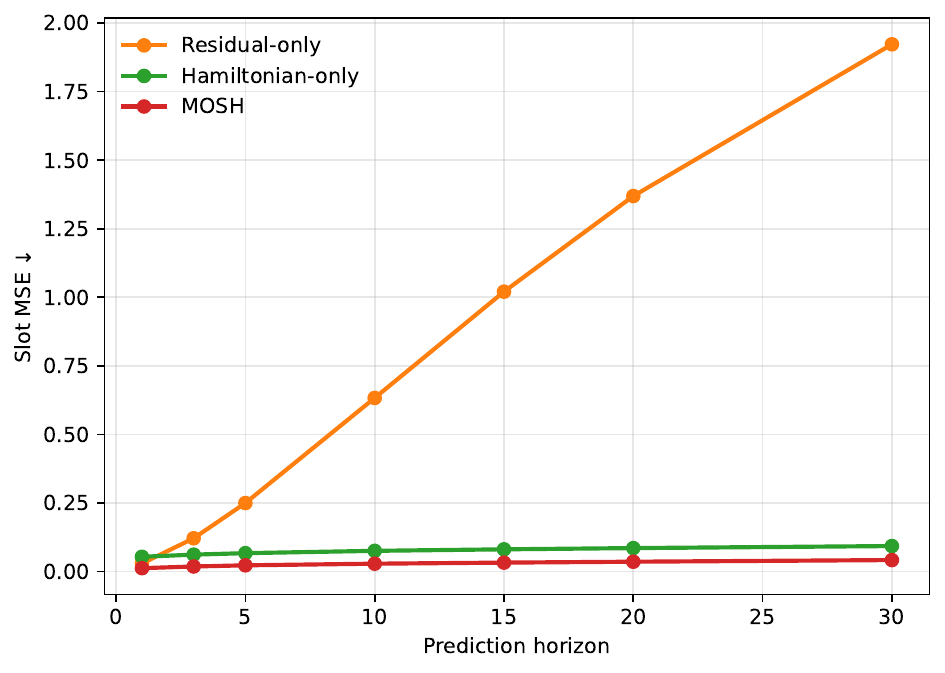}
\caption{Physion transfer through frozen STEVE. Slot MSE remains lowest for the complete model throughout the 35-step closed-loop rollout.}
\label{fig:physion-summary}
\end{figure}

\paragraph{Physion transfer.}
Figure~\ref{fig:physion-summary} evaluates a separately trained STEVE interface. Full MOSH is much more stable than residual-only prediction and improves over Hamiltonian-only prediction across the rollout, supporting the combined physics--appearance pathway across observers. Scenario results and manifests are in the supplement, completing the transfer evidence after OBJ3D, CLEVRER, and the long-horizon OBJ3D ablations.
The Physion result is primarily a representation-transfer test: STEVE differs from SAVi in slot dimension, resolution, and decoder behavior, so slot MSE is the meaningful cross-observer metric. The complete model remains below both reduced variants throughout the rollout, supporting the intended division of labor between phase dynamics and contextual appearance. The scenario breakdown uses the same locked no-feedback protocol, so differences reflect interaction difficulty rather than changes in training.

The horizon-resolved values further quantify this transfer. Full MOSH records slot MSE of $0.0126$, $0.0187$, $0.0230$, $0.0290$, $0.0362$, and $0.0422$ at horizons 1, 3, 5, 10, 20, and 30, respectively. Hamiltonian-only prediction remains bounded but is consistently higher, reaching $0.0932$ at horizon 30, whereas residual-only prediction accumulates rapidly to $1.9220$. The gap is already visible at horizon 3 and widens throughout the rollout, indicating that the contextual branch is not merely repairing the first predicted frame; it helps preserve a stable decoder-compatible trajectory over repeated feedback.
Scenario-level results show the same pattern across different interactions. At horizon 30, the lowest slot error occurs on Support ($0.0278$) and Drop ($0.0321$), while Drape ($0.0683$) and Collide ($0.0546$) are more difficult. These differences are expected from the changing contact and occlusion patterns, but the common evaluator and frozen observer make the breakdown useful as a robustness check. Together with the aggregate curve, the scenario results support transfer of the phase/context factorization rather than performance on a single favorable physical setting.

\enlargethispage{8\baselineskip}
\section{Discussion and Limitations}
On OBJ3D, the complete model leads in image and observer space, its margin grows under feedback, and each component removal increases error. The experimental story is therefore coherent: the interface is faithful, the benchmark gains are broad, the long-horizon drift is smaller, and the transfer result holds on a different frozen observer.

\noindent\textbf{Limitations.} Results are single validation-selected runs. The state inherits observer errors, and its energy is a soft bias rather than an identified conserved Hamiltonian. PredRNN lacks aligned slots, while CLEVRER and Physion are fixed-compute transfers with protocol-specific metrics. We study passive rollout rather than control or planning.

\section{Conclusion}
\noindent\textbf{Summary.} \method{} separates mask-grounded phase state from decoder-oriented context. Slot support defines $Q$, temporal differences define $P$, soft Hamiltonian guidance structures rollout, and a causal composer restores decoder-relevant appearance.

\noindent\textbf{Evidence.} On 6-to-30 OBJ3D, LPIPS falls by 25.0\% and spatial MSE by 33.7\% versus the strongest object-centric baseline; object-state metrics, horizon curves, and ablations agree. CLEVRER transfers favorably, while Physion confirms the combined pathway under a distinct STEVE interface.

\noindent\textbf{Future work.} We will improve phase extraction under occlusion and add interaction, depth, and action variables.

\clearpage
\bibliography{aaai2027}

\clearpage
\appendix
\input{AppendixBody.tex}
\end{document}

%% file: AppendixBody.tex
\section{Additional Method Details}
\subsection{Additional Overview}
Given $L$ observed frames $X_{1:L}$, the frozen \savi{} interface returns visual slots $S_{1:L}$ and soft masks $M_{1:L}$. \method{} maintains a canonical phase state $(Q_t,P_t)$ and a causal visual context $C_t$. The phase state is used exclusively by the Soft-Hamiltonian transition, whereas $C_t$ retains appearance and identity information for decoding.

\subsection{Mask-Grounded Phase-State Factorization}
Slots are optimized for grouping and reconstruction and may therefore entangle geometry with appearance and semantics \cite{slotattention,kipf2022savi}. We retain the support owned by slot $n$ through winner-take-all assignment, $M^{\mathrm{own}}_{t,n}(u,v)=M_{t,n}(u,v)\mathbb I[n=\arg\max_j M_{t,j}(u,v)]$. Let $b$ be a fixed spatial basis and let $\operatorname{Pool}$ be normalized mask-weighted pooling. The position-like coordinate is
\begin{equation}
Q_{t,n}=f_Q\!\left(\operatorname{Pool}(M^{\mathrm{own}}_{t,n},b)\right).
\label{app-eq:qstate}
\end{equation}
The lightweight map $f_Q$ contains a basis projector, a slot embedding, and one attention-MLP block. Its inputs are mask support and fixed spatial bases, which separate the canonical coordinate from direct slot appearance. Its momentum-like companion is the temporal finite difference,
\begin{equation}
P_t=\frac{Q_t-Q_{t-1}}{\Delta t}.
\label{app-eq:pstate}
\end{equation}
In parallel, a causal spatiotemporal encoder reads slot history and produces $C_t=\mathcal F_{\mathrm{ctx}}(S_{\leq t})$. Context never enters the phase-state extractor; it is reserved for the decoder-compatible pathway below.

\subsection{Soft-Hamiltonian Closed-Loop Dynamics}
A fully learned transition can fit local visual correlations without a preferred physical direction, while a strict Hamiltonian model can be too restrictive for imperfect visual observations. We therefore use a soft physical bias. The energy head receives only the canonical pair,
\begin{equation}
H_t=H_{\vartheta}(Q_t,P_t).
\label{app-eq:energy}
\end{equation}
A learned core conditioned on $[Q_t,P_t]$ and $C_t$ predicts increments $(D_t^Q,D_t^P)$. We blend these increments with the energy-gradient directions:
\begin{align}
Q_{t+1}&=Q_t+(1-\alpha)D_t^Q+\alpha\operatorname{clip}_{\gamma}(\nabla_{P_t}H_t),\\
P_{t+1}&=P_t+(1-\alpha)D_t^P-\alpha\operatorname{clip}_{\gamma}(\nabla_{Q_t}H_t).
\label{app-eq:update}
\end{align}
The learned path captures residual dynamics, whereas the gradient path organizes phase evolution. These are forward-Euler Soft-Hamiltonian updates, not a symplectic solver or a claim that $H_t$ is ground-truth mechanical energy. During rollout, the model directly sets $(Q_t,P_t)\leftarrow(Q_{t+1},P_{t+1})$.

\subsection{Physics--Appearance Reconciliation}
The phase state intentionally omits decoder-relevant visual information. After the physical update, we use a gated composer to combine the normalized phase displacement with $C_t$, yielding a physical slot proposal $S_{t+1}^{\mathrm{phys}}$. A causal reasoning module then provides a small decoder-space correction:
\begin{equation}
\widehat S_{t+1}=S_{t+1}^{\mathrm{phys}}+\beta\bigl(\mathcal R(S_t,S_{\leq t})-S_t\bigr).
\label{app-eq:reconcile}
\end{equation}
The composer applies learned projections to normalized phase displacement and $C_t$, predicts a sigmoid gate, and blends the two projected features before mapping them to slot space. The frozen decoder produces $\widehat X_{t+1}=\mathcal D(\widehat S_{t+1})$, and $\widehat S_{t+1}$ is appended to the history used for $C_{t+1}$.

\subsection{Learning Objective}
We freeze the \savi{} encoder and decoder, encode $L=6$ frames, and unroll $K$ future steps according to the training curriculum. The complete objective is
\begin{align}
\mathcal L={}&\mathcal L_{\mathrm{slot}}+\mathcal L_{\mathrm{img}}+\lambda_{\mathrm{ham}}\mathcal L_{\mathrm{ham}}+\lambda_{\mathrm{energy}}\mathcal L_{\mathrm{energy}}\\
&+\lambda_{\mathrm{ctr}}\mathcal L_{\mathrm{ctr}}+\lambda_{\mathrm{res}}\mathcal L_{\mathrm{res}}.
\end{align}
The image term combines pixel fidelity, pooled multiscale fidelity, and gradient matching. The Hamiltonian term aligns the learned increments with the energy gradient, the energy term encourages rollout consistency, the centroid term stabilizes geometry, and the residual term constrains the decoder-side correction. During the early phase of training, the physical coefficients are linearly warmed up so appearance reconstruction stabilizes first.

\section{Training Diagnostic}
The main OBJ3D run uses the shared configuration reported in the experimental setup. The validation curve below documents checkpoint selection without repeating the protocol details.

\begin{figure}[H]
\centering
\includegraphics[width=\columnwidth]{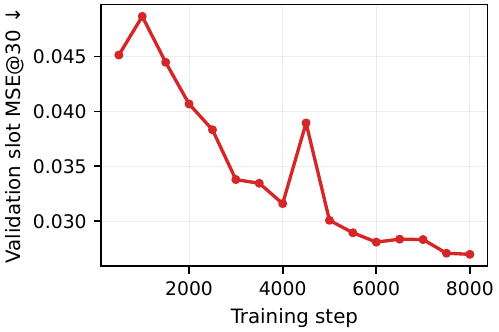}
\caption{Validation Slot@30 during the main OBJ3D training run. The curve is used for checkpoint selection.}
\label{app-fig:training-curve}
\end{figure}

We vary one architectural or functional component at a time under a shared short-budget protocol. Shape-compatible tensors inherit the validation-selected full-model initialization; changed shapes are reinitialized. These experiments measure directional sensitivity rather than multi-seed confidence intervals.

\begin{table}[H]
\centering
\small
\setlength{\tabcolsep}{3.2pt}
\begin{tabular}{@{}lrrr@{}}
\toprule
Attention blocks & LPIPS $\downarrow$ & MSE $\downarrow$ & Slot@30 $\downarrow$\\
\midrule
1 & 0.1863 & 9.945 & 0.06165\\
2 (default) & \textbf{0.0873} & \textbf{4.789} & \textbf{0.02755}\\
3 & 0.1251 & 6.856 & 0.03984\\
\bottomrule
\end{tabular}

\vspace{4pt}
\begin{tabular}{@{}lrrr@{}}
\toprule
MLP ratio & LPIPS $\downarrow$ & MSE $\downarrow$ & Slot@30 $\downarrow$\\
\midrule
1 & 0.2662 & 12.266 & 0.07734\\
2 (default) & \textbf{0.0873} & \textbf{4.789} & \textbf{0.02755}\\
4 & 0.2193 & 10.887 & 0.06511\\
\bottomrule
\end{tabular}
\caption{Capacity sensitivity under the shared short-budget protocol.}
\label{app-tab:capacity-ablation}
\end{table}

The complete model is preferred in both capacity sweeps. The two-block configuration balances expressivity and rollout stability, while the default MLP ratio avoids the degradation observed at narrower and wider settings.

\section{Additional OBJ3D Results}
\subsection{Independent Long-Horizon Rollout}
The following sequence is independent of the main-paper qualitative strip. All methods receive the same six observations and generate the next 30 frames autoregressively. The visualization emphasizes identity retention, relative layout, and the growth of errors under repeated feedback.

\begin{figure}[H]
\centering
\includegraphics[width=\columnwidth,height=.43\textheight,keepaspectratio]{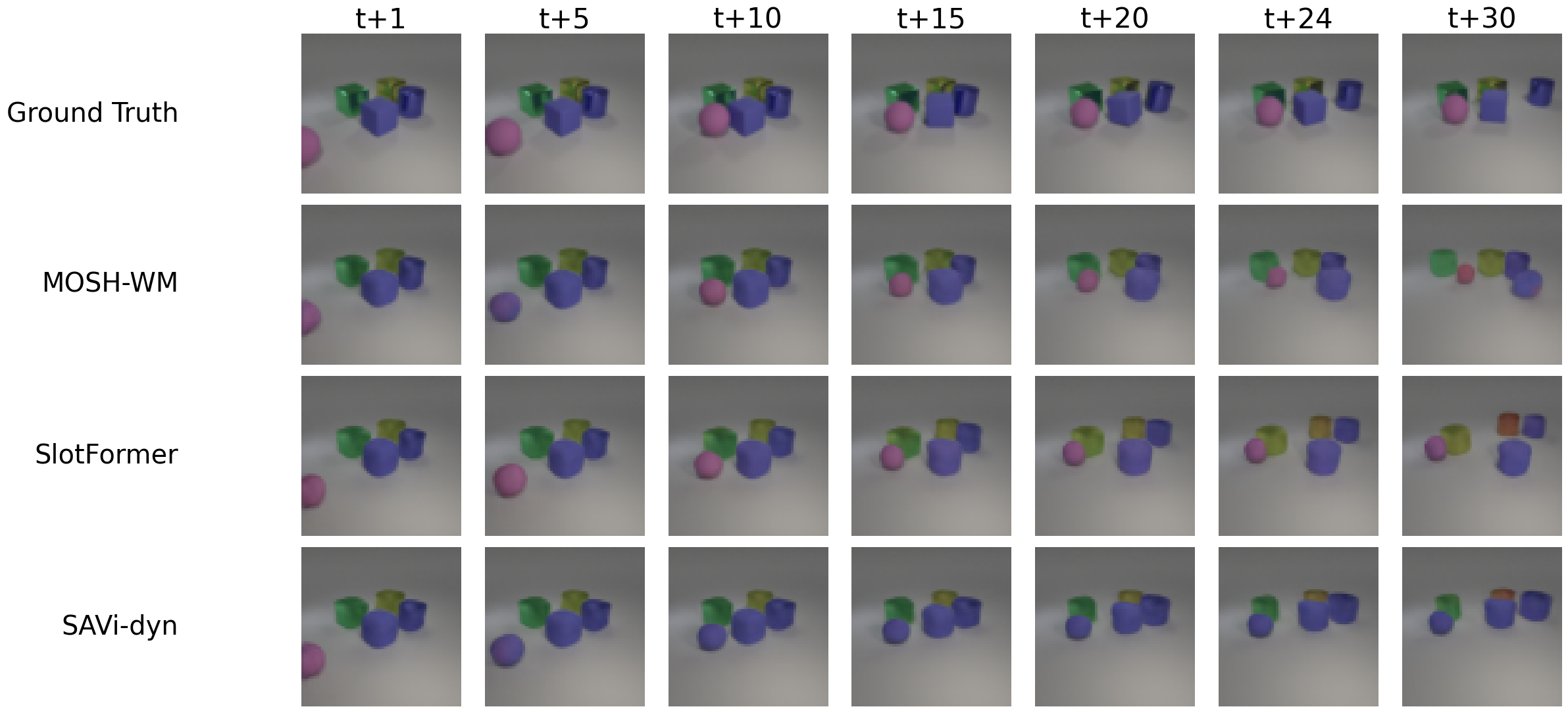}
\caption{Additional OBJ3D 6-to-30 rollout on test video 125. Columns show future horizons and rows compare ground truth with three predictors under the same closed-loop protocol.}
\label{app-fig:additional-rollout}
\end{figure}

\subsection{Error and Transfer Summaries}
The compact plots below are additional summaries rather than replacements for the main-paper curves. The OBJ3D plot reports relative degradation when functional pathways are removed. The Physion plot evaluates the same phase/context factorization through a different frozen observer.
\begin{figure}[H]
\centering
\includegraphics[width=.72\columnwidth]{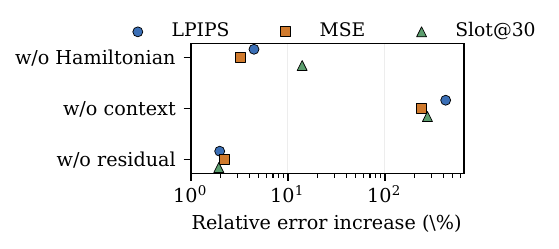}
\caption{Compact OBJ3D ablation summary. The plot shows relative error increases when the physical, context, or residual pathway is removed.}
\label{app-fig:appendix-horizon-compact}
\end{figure}

\begin{figure}[H]
\centering
\includegraphics[width=.86\columnwidth]{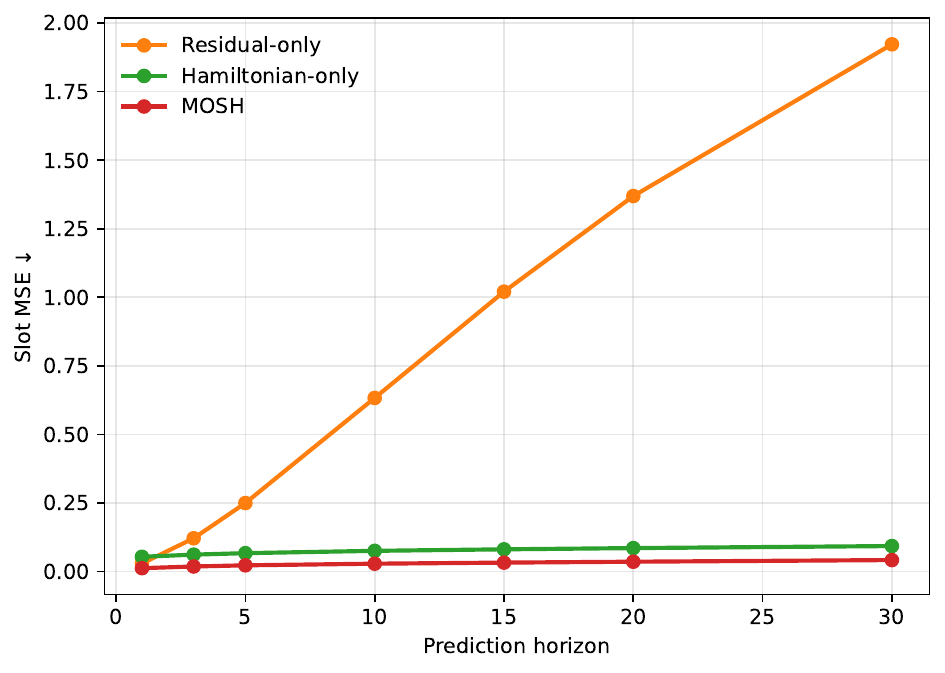}
\caption{Physion slot error over the prediction horizon. The complete model remains lowest throughout the rollout.}
\label{app-fig:appendix-physion-curve}
\end{figure}
\section{Inference Procedure}
For completeness, one prediction step can be written as a fixed sequence of interface-preserving operations. First, the frozen observer encodes the current frame and returns slots and masks. Second, owned-mask pooling produces the canonical coordinate for each slot, and the temporal difference produces the momentum-like variable. Third, the learned core and the energy head jointly produce the phase update. Fourth, the composer combines the updated phase state with the causal context and applies the bounded decoder-space correction. Finally, the frozen decoder reconstructs the next frame and the predicted slots are appended to the context history.

\begin{equation}
\begin{split}
(S_t,M_t)&=\mathcal E(X_t),\\
Q_t&=f_Q\!\left(\operatorname{Pool}(M_t^{\mathrm{own}},b)\right),\\
P_t&=(Q_t-Q_{t-1})/\Delta t,\\
(Q_{t+1},P_{t+1})&=\Phi_{\vartheta}(Q_t,P_t,C_t),\\
\widehat S_{t+1}&=\Psi(Q_{t+1},P_{t+1},C_t,S_t),\\
\widehat X_{t+1}&=\mathcal D(\widehat S_{t+1}).
\end{split}
\label{app-eq:inference-chain}
\end{equation}

The same chain is used during teacher-forced training and closed-loop evaluation, except that training may use the target interface slots in the loss computation. During evaluation, the predicted frame and predicted slots are fed back without target feedback. This distinction keeps the evaluator identical across methods while exposing accumulated rollout error.

\subsection{Temporal Weighting and Rollout Stability}
Let $e_k$ denote an observer-space error at future step $k$. The rollout objective aggregates errors across the prediction window using nondecreasing temporal weights $w_k$:
\begin{equation}
\mathcal L_{\mathrm{roll}}=\frac{1}{\sum_{k=1}^{K}w_k}\sum_{k=1}^{K}w_k e_k,
\qquad w_{k+1}\geq w_k>0.
\label{app-eq:temporal-loss}
\end{equation}
This weighting does not change the evaluator or the reported metrics. It only makes late rollout errors visible during optimization. The energy and centroid terms provide complementary constraints: the former discourages abrupt phase inconsistency, while the latter directly penalizes errors in the spatial support summarized by the masks.

A useful diagnostic is the ratio between the decoder correction and the composed physical proposal,
\begin{equation}
\rho_t=\frac{\lVert r_{t+1}\rVert_2}{\epsilon+\lVert S_{t+1}^{\mathrm{phys}}\rVert_2},
\label{app-eq:residual-ratio}
\end{equation}
where $\epsilon$ is a numerical stabilizer. The residual regularizer acts on this ratio rather than allowing the correction to replace the phase-driven proposal. This is why the residual branch can restore decoder-relevant details while the phase branch remains responsible for the propagated state.

\subsection{Reproducibility Notes}
All methods use the same observation window, frozen observer, decoder, image normalization, and closed-loop evaluator for each benchmark. Model selection is performed on validation data using the specified horizon-resolved slot metric. Test frames are used only for the final reported evaluation and qualitative exports. The files in \texttt{physion\_results/} provide the transfer tables and manifests used to produce the Physion results shown above.

The appendix figures are intentionally separated into three roles. The training curve documents optimization and checkpoint selection; the additional OBJ3D strip tests visual stability on a sequence not used in the main qualitative figure; and the Physion visualization demonstrates transfer through a different frozen slot interface. This separation avoids counting the same visualization as independent evidence while making the evaluation protocol explicit.

\section{Symbols and Tensor Shapes}
For clarity, the main tensors used by the implementation are summarized below. The number of slots is denoted by $N$, the slot dimension by $d_s$, the phase dimension by $d_q$, and the number of observed frames by $L$. The observer returns $S_t\in\mathbb R^{N\times d_s}$ and $M_t\in\mathbb R^{N\times H\times W}$; the phase extractor returns $Q_t,P_t\in\mathbb R^{N\times d_q}$; and the causal context retains a history of decoder-compatible slot tokens. The same slot ordering is used for mask ownership, phase extraction, composition, and observer-space evaluation.

\begin{table}[H]
\centering
\scriptsize
\setlength{\tabcolsep}{3pt}
\begin{tabular}{@{}lll@{}}
\toprule
Symbol & Shape & Meaning\\
\midrule
$X_t$ & $H\times W\times 3$ & video frame\\
$S_t$ & $N\times d_s$ & observer slot tokens\\
$M_t$ & $N\times H\times W$ & soft slot masks\\
$Q_t$ & $N\times d_q$ & mask-grounded coordinate\\
$P_t$ & $N\times d_q$ & temporal phase difference\\
$C_t$ & context-dependent & causal visual memory\\
$\widehat S_t$ & $N\times d_s$ & decoder-compatible prediction\\
\bottomrule
\end{tabular}
\caption{Notation and tensor roles used in the implementation.}
\label{app-tab:symbols}
\end{table}

\subsection{Implementation-Level Update}
The update is implemented slot-wise after the shared interaction blocks. Let $z_t=[Q_t,P_t,C_t]$ denote the concatenated state available to the learned core. The core produces a bounded proposal $D_t=\mathcal F_\theta(z_t)$, while the energy head produces the directional term $G_t=(\nabla_{P_t}H_t,-\nabla_{Q_t}H_t)$. The actual phase increment is therefore
\begin{equation}
\Delta_t=(1-\alpha)D_t+\alpha\operatorname{clip}_{\gamma}(G_t),
\qquad
[Q_{t+1},P_{t+1}]=[Q_t,P_t]+\Delta_t.
\label{app-eq:implementation-update}
\end{equation}
The clipping operator is applied component-wise after the energy gradient is computed. It limits an occasional high-magnitude gradient without changing the direction of ordinary updates. The learned proposal remains available even when the energy signal is weak, which is the reason the method is described as soft-Hamiltonian rather than strictly Hamiltonian.

The decoder-side composer is evaluated after the phase update. Its gate is computed from both phase and context, so the model can use phase information to determine where motion is expected while using context to preserve details that are not encoded in the canonical state. This separation also explains why the same phase state can be evaluated through different frozen observers: the observer-specific decoder interface is handled after the shared transition logic.

\subsection{Closed-Loop Pseudocode}
The following compact procedure describes the evaluation-time rollout without introducing an additional training assumption.

\begin{quote}
\small
\textbf{Closed-loop rollout.} Encode the observed frames with the frozen observer. Initialize $Q_L$ from the owned masks, $P_L$ from the last temporal difference, and $C_L$ from the observed slot history. For $k=1,\ldots,K$: compute $H_L$ and its gradients; blend the learned and gradient directions; update $Q$ and $P$; compose the next decoder-compatible slots; decode the predicted frame; and append the predicted slots to the causal context. Return all predicted frames and observer-space states.
\end{quote}

This procedure keeps the feedback path explicit. The target future frames are used only to compute evaluation losses; they are never used to update $Q$, $P$, or $C$ during closed-loop testing. Consequently, the horizon curves measure accumulated error rather than one-step reconstruction quality.

\section{Physion Transfer}
We follow the SlotFormer Physion protocol. Videos are truncated to 150 source frames and sampled every three frames; STEVE produces six 192-dimensional slots at $128\times128$ resolution. MOSH observes 15 sampled frames, trains with a 10-step closed-loop rollout, and is evaluated for the remaining 35 sampled steps without target feedback.

\begin{table}[H]
\centering
\small
\setlength{\tabcolsep}{2.6pt}
\begin{tabular}{@{}lrrrrrr@{}}
\toprule
Method & @1 & @3 & @5 & @10 & @20 & @30\\
\midrule
Residual-only & .0296 & .1215 & .2500 & .6331 & 1.3689 & 1.9220\\
Hamiltonian-only & .0541 & .0620 & .0674 & .0758 & .0858 & .0932\\
MOSH (full) & \textbf{.0126} & \textbf{.0187} & \textbf{.0230} & \textbf{.0290} & \textbf{.0362} & \textbf{.0422}\\
\bottomrule
\end{tabular}
\caption{Physion test slot MSE over the closed-loop rollout.}
\label{app-tab:appendix-physion}
\end{table}

\begin{figure}[H]
\centering
\includegraphics[width=.92\columnwidth]{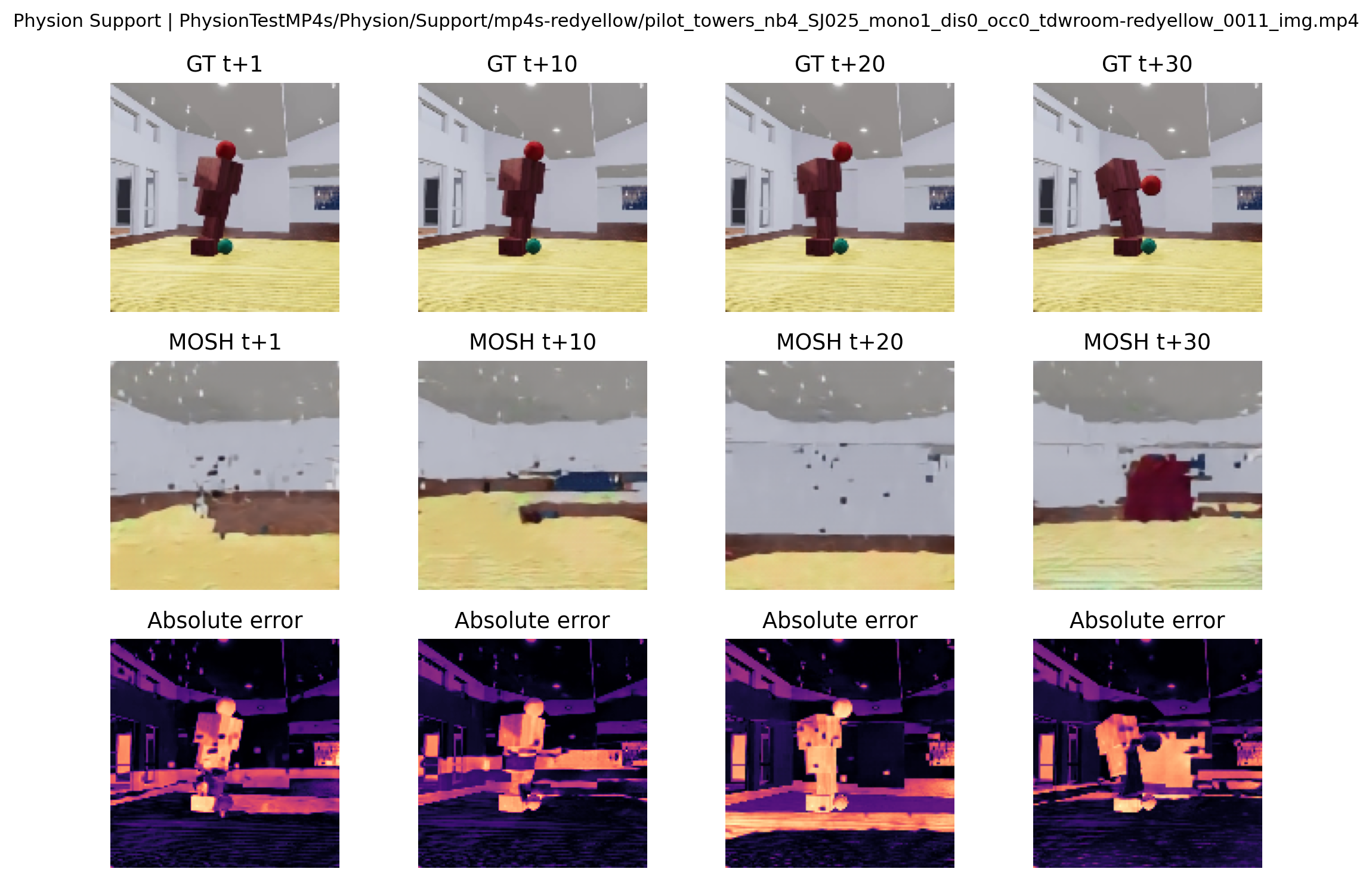}
\caption{Physion qualitative transfer. Ground truth, MOSH prediction, and absolute reconstruction error are shown at horizons 1, 10, 20, and 30. Pixel quality is limited by the frozen STEVE decoder; the experiment is primarily a slot-dynamics transfer test.}
\label{app-fig:appendix-physion-qualitative}
\end{figure}

\begin{table}[H]
\centering
\scriptsize
\begin{tabular}{@{}lr@{}}
\toprule
Scenario & MOSH @30\\
\midrule
Collide & 0.0546\\ Contain & 0.0341\\ Dominoes & 0.0403\\ Drape & 0.0683\\
Drop & 0.0321\\ Link & 0.0362\\ Roll & 0.0437\\ Support & 0.0278\\
\bottomrule
\end{tabular}
\caption{Scenario-level Physion slot MSE at horizon 30.}
\label{app-tab:appendix-scenarios}
\end{table}

\begin{table}[H]
\centering
\small
\begin{tabular}{@{}ll@{}}
\toprule
Item & Setting\\
\midrule
Observer & frozen SAVi / STEVE\\
OBJ3D horizon & 6 observed, 30 predicted\\
Optimizer & Adam, $10^{-5}$\\
Seed & 42\\
Selection & validation Slot@30\\
Feedback & disabled\\
\bottomrule
\end{tabular}
\caption{Reproducibility checklist.}
\label{app-tab:appendix-checklist}
\end{table}

The local observer and world-model manifests, result tables, and scenario breakdowns are included in \texttt{physion\_results/}. All reported runs use seed 42 and the same evaluator; no target feedback is used during test rollout.
\subsection{Interpretation of the Transfer Result}
The Physion experiment is designed to test whether the phase/context factorization survives a change in the upstream observer. The STEVE interface differs from the SAVi interface in slot dimension, image resolution, and decoder behavior, so the absolute pixel appearance is not directly comparable to OBJ3D. The reported slot metric is therefore the primary transfer measurement. The full model remains below the two reduced variants throughout the listed horizons, which is consistent with the intended division of labor: the phase branch carries the evolving spatial state, while the contextual branch preserves information required by the downstream decoder.

The scenario table should be read as a breakdown of the same locked test protocol rather than as a new training set. Each scenario uses the same observation length, rollout rule, and no-feedback evaluation. Differences between scenarios reflect the difficulty of the physical interaction, while the aggregate curve summarizes the common trend across the transfer suite. We avoid interpreting the absolute reconstruction error as a standalone image-quality claim because the frozen STEVE decoder is not optimized for pixel-perfect reconstruction in this setting.

%% file: aaai2027.bib
@inproceedings{slotattention,author={Locatello, Francesco and others},title={Object-Centric Learning with Slot Attention},booktitle={NeurIPS},year={2020}}

@inproceedings{monet,author={Burgess, Christopher P. and others},title={MONet: Unsupervised Scene Decomposition and Representation},booktitle={ICML},year={2019}}

@inproceedings{iodine,author={Greff, Klaus and others},title={Multi-Object Representation Learning with Iterative Variational Inference},booktitle={ICLR},year={2019}}

@inproceedings{kipf2022savi,author={Kipf, Thomas and others},title={Conditional Object-Centric Learning from Video},booktitle={ICLR},year={2022}}

@inproceedings{slotformer,author={Wu, Ziyi and others},title={SlotFormer: Unsupervised Visual Dynamics Simulation with Object-Centric Models},booktitle={ICLR},year={2023}}

@inproceedings{gswm,author={Lin, Zhixuan and others},title={Improving Generative Imagination in Object-Centric World Models},booktitle={ICML},year={2020},pages={6140--6149}}

@inproceedings{interactionnetworks,author={Battaglia, Peter W. and others},title={Interaction Networks for Learning about Objects, Relations and Physics},booktitle={NeurIPS},year={2016}}

@inproceedings{nri,author={Kipf, Thomas N. and others},title={Neural Relational Inference for Interacting Systems},booktitle={ICML},year={2018}}

@inproceedings{greydanus2019hamiltonian,author={Greydanus, Samuel and others},title={Hamiltonian Neural Networks},booktitle={NeurIPS},year={2019}}

@inproceedings{cranmer2020lagrangian,author={Cranmer, Miles and others},title={Lagrangian Neural Networks},booktitle={ICLR},year={2020}}

@inproceedings{slotpi,author={Li, Jian and others},title={SlotPi: Physics-informed Object-centric Reasoning Models},booktitle={ACM Multimedia},year={2025}}

@article{hamworld,author={Tang, Haoyun and others},title={HaM-World: Soft-Hamiltonian World Models with Selective Memory for Planning},journal={arXiv preprint arXiv:2605.05951},year={2026}}

@inproceedings{clevrer,author={Yi, Kexin and others},title={CLEVRER: Collision Events for Video Representation and Reasoning},booktitle={ICLR},year={2020}}

@inproceedings{phyre,author={Bakhtin, Anton and others},title={PHYRE: A New Benchmark for Physical Reasoning},booktitle={NeurIPS},year={2019}}

@inproceedings{lpips,author={Zhang, Richard and others},title={The Unreasonable Effectiveness of Deep Features as a Perceptual Metric},booktitle={CVPR},year={2018}}

@article{ssim,author={Wang, Zhou and others},title={Image Quality Assessment: From Error Visibility to Structural Similarity},journal={IEEE TIP},year={2004}}

@inproceedings{air,author={Eslami, S. M. Ali and others},title={Attend, Infer, Repeat: Fast Scene Understanding with Generative Models},booktitle={NeurIPS},year={2016}}

@inproceedings{space,author={Lin, Zhixuan and others},title={SPACE: Unsupervised Object-Oriented Scene Representation via Spatial Attention and Decomposition},booktitle={ICLR},year={2020}}

@inproceedings{genesis,author={Engelcke, Martin and others},title={GENESIS: Generative Scene Inference and Sampling with Object-Centric Latent Representations},booktitle={ICLR},year={2020}}

@inproceedings{genesisv2,author={Engelcke, Martin and others},title={GENESIS-V2: Inferring Unordered Object-Centric Scene Representations without Iterative Refinement},booktitle={NeurIPS},year={2021}}

@inproceedings{scalor,author={Jiang, Jindong and others},title={SCALOR: Generative World Models with Scalable Object-Oriented Representations},booktitle={ICLR},year={2020}}

@inproceedings{op3,author={Veerapaneni, Rishabh and others},title={Entity Abstraction in Visual Model-Based Reinforcement Learning},booktitle={CoRL},year={2020}}

@inproceedings{cswm,author={Kipf, Thomas and others},title={Contrastive Learning of Structured World Models},booktitle={ICLR},year={2020}}

@inproceedings{steve,author={Singh, Gautam and others},title={STEVE: Slot Transformer for Object-Centric Learning},booktitle={NeurIPS},year={2022}}

@inproceedings{dinosaur,author={Seitzer, Maximilian and others},title={Bridging the Gap to Real-World Object-Centric Learning},booktitle={ICLR},year={2023}}

@inproceedings{vin,author={Watters, Nicholas and others},title={Visual Interaction Networks: Learning a Physics Simulator from Video in the Wild},booktitle={NeurIPS},year={2017}}

@inproceedings{npe,author={Chang, Michael B. and others},title={A Compositional Object-Based Approach to Learning Physical Dynamics},booktitle={ICLR},year={2017}}

@inproceedings{gns,author={Sanchez-Gonzalez, Alvaro and others},title={Learning to Simulate Complex Physics with Graph Networks},booktitle={ICML},year={2020}}

@article{gnns,author={Battaglia, Peter W. and others},title={Relational Inductive Biases, Deep Learning, and Graph Networks},journal={arXiv preprint arXiv:1806.01261},year={2018}}

@inproceedings{ode,author={Chen, Ricky T. Q. and others},title={Neural Ordinary Differential Equations},booktitle={NeurIPS},year={2018}}

@inproceedings{latentode,author={Rubanova, Yulia and others},title={Latent Ordinary Differential Equations for Irregularly-Sampled Time Series},booktitle={NeurIPS},year={2019}}

@inproceedings{symplectic,author={Zhong, Yaofeng and others},title={Symplectic ODE-Net: Learning Hamiltonian Dynamics with Control},booktitle={ICLR},year={2020}}

@inproceedings{hgn,author={Toth, Peter and others},title={Hamiltonian Generative Networks},booktitle={ICLR},year={2020}}

@inproceedings{deeplagrangian,author={Lutter, Michael and others},title={Deep Lagrangian Networks: Using Physics as Model Prior for Deep Learning},booktitle={ICLR},year={2019}}

@inproceedings{vintegrator,author={Toth, Peter and others},title={Hamiltonian Generative Networks with Nonseparable Hamiltonians},booktitle={NeurIPS},year={2020}}

@inproceedings{worldmodels,author={Ha, David and Schmidhuber, Jurgen},title={World Models},booktitle={NeurIPS Workshop},year={2018}}

@inproceedings{planet,author={Hafner, Danijar and others},title={Learning Latent Dynamics for Planning from Pixels},booktitle={ICML},year={2019}}

@inproceedings{dreamer,author={Hafner, Danijar and others},title={Dream to Control: Learning Behaviors by Latent Imagination},booktitle={ICLR},year={2020}}

@inproceedings{dreamerv2,author={Hafner, Danijar and others},title={Mastering Atari with Discrete World Models},booktitle={ICLR},year={2021}}

@article{muzero,author={Schrittwieser, Julian and others},title={Mastering Atari, Go, Chess and Shogi by Planning with a Learned Model},journal={Nature},volume={588},pages={604--609},year={2020}}

@inproceedings{physion,author={Bear, Daniel M. and others},title={Physion: Evaluating Physical Prediction from Vision in Humans and Machines},booktitle={NeurIPS},year={2021}}

@inproceedings{clevr,author={Johnson, Justin and others},title={CLEVR: A Diagnostic Dataset for Compositional Language and Elementary Visual Reasoning},booktitle={CVPR},year={2017}}

@inproceedings{rpin,author={Xu, Zhenjia and others},title={Unsupervised Discovery of Parts, Structure, and Dynamics},booktitle={ICLR},year={2019}}

@inproceedings{transformer,author={Vaswani, Ashish and others},title={Attention Is All You Need},booktitle={NeurIPS},year={2017}}

@inproceedings{perceiver,author={Jaegle, Andrew and others},title={Perceiver: General Perception with Iterative Attention},booktitle={ICML},year={2021}}

@inproceedings{slotmixer,author={Lippe, Phillip and others},title={Efficient Object-Centric Learning with Slot Mixing},booktitle={NeurIPS},year={2023}}

@inproceedings{ocvp,author={Sajjadi, Mehdi and others},title={Object-Centric Video Prediction},booktitle={NeurIPS},year={2022}}

@inproceedings{tsp,author={Denton, Emily and Fergus, Rob},title={Stochastic Video Generation with a Learned Prior},booktitle={ICML},year={2018}}

@inproceedings{svg,author={Villegas, Ruben and others},title={Decomposing Motion and Content for Natural Video Sequence Prediction},booktitle={ICLR},year={2017}}

@inproceedings{predrnn,author={Wang, Yunbo and others},title={PredRNN: A Recurrent Neural Network for Spatiotemporal Predictive Learning},booktitle={NeurIPS},year={2017}}

@inproceedings{simvp,author={Gao, Zhangyang and others},title={SimVP: Simpler Yet Better Video Prediction},booktitle={CVPR},year={2022},pages={3170--3180}}

@article{gqn,author={Eslami, S. M. Ali and others},title={Neural Scene Representation and Rendering},journal={Science},volume={360},number={6394},pages={1204--1210},year={2018}}
